# Bi-CamoDiffusion: A Boundary-informed Diffusion Approach for Camouflaged Object Detection

Patricia L. Suárez[1] Leo Thomas Ramos[2,3] Angel D. Sappa[1,2,3]
[1]ESPOL Polytechnic University [2]Computer Vision Center [3]Universitat Autònoma de Barcelona
plsuarez@espol.edu.ec; ltramos@cvc.uab.cat; sappa@ieee.org

## Abstract

*Bi-CamoDiffusion is introduced, an evolution of the CamoDiffusion framework for camouflaged object detection. It integrates edge priors into early-stage embeddings via a parameter-free injection process, enhancing boundary sharpness and preventing structural ambiguity. An optimization objective that unifies spatial accuracy, structural constraints, and uncertainty supervision is also proposed, allowing the model to capture of both the object's global context and its intricate boundary transitions. Evaluations across the CAMO, COD10K, and NC4K datasets show that Bi-CamoDiffusion surpasses the baseline, delivering sharper delineation of thin structures and protrusions while also minimizing false positives. The model consistently outperforms existing state-of-the-art methods across all evaluated metrics, including $S_m$, $F^w_\beta$, $E_m$, and $MAE$, demonstrating a more precise object-background separation and sharper boundary recovery. Code available at https://github.com/plsuarez/Bi-CamoDiffusion*

## 1. Introduction

Camouflaged Object Detection (COD) involves the pixel-level localization of objects whose appearance exhibits extremely low foreground-background separability [8, 11]. The objective is to generate accurate segmentation masks under these conditions [25], enabling reliable object delineation even when visual evidence is weak. Unlike conventional segmentation settings that rely on dominant visual cues [23], COD addresses scenarios in which object presence must be inferred from subtle structural and contextual signals [25, 35].

COD is challenging due to multiple factors, including intrinsic target-background similarity [8], large variations in object scale [1], and frequent occlusions disrupting spatial continuity [7]. Beyond these, a key challenge lies in the reliable delineation of object boundaries [7, 37]. When visual transitions are weak or fragmented, determining where the object ends becomes uncertain, often leading to unstable masks and imprecise spatial support [27]. Consequently, boundary ambiguity emerges as a central obstacle in accurate COD.

Diverse strategies have been proposed for COD, including Convolutional Neural Networks (CNNs), attention-driven models [13, 40], and more recently, generative frameworks [13]. Among these, diffusion models have gained particular relevance by formulating the prediction as an iterative denoising process [10, 24, 26]. However, diffusion models present certain limitations, as they do not explicitly enforce structural constraints [17] during mask formation. Consequently, predictions may produce plausible regions but fail to preserve rigid geometric structures [32], a deficiency that becomes critical when precise boundary delineation is required.

Based on the above, and building upon CamoDiffusion [2], an structurally informed extension designed to improve boundary delineation is introduced. The proposed approach incorporates structural cues directly into the early feature space, enabling the denoising process to better preserve spatial support as the mask progressively emerges. This reinforces contour fidelity in visually ambiguous regions. The approach preserves the generative process without altering the underlying backbone, maintaining architectural compatibility.

To further support boundary delineation, a multi-scale boundary-informed objective is proposed. The loss extends the base diffusion segmentation supervision with three complementary four applied across output scales: a focal structure component that emphasizes hard pixels while preserving region level structure, a boundary term that enforces contour agreement between predictions and ground-truths, and an uncertainty aware regularization term that penalizes ambiguous responses to stabilize optimization. Also, an RGB boundary consistency constraint aligns predicted contours with structural cues derived from the input image, reducing boundary bleeding.

Experiments on the CAMO, COD10K, and NC4K datasets demonstrate that the proposed approach achieves consistent improvements in detection quality, producing sharper contours and more reliable predictions in challenging camouflaged scenes. The main contributions of this work are summarized as follows:

- A boundary-informed extension of CamoDiffusion with edge-guided early feature injection and no additional learnable parameters.
- A multi-scale training objective for contour-aware and structurally consistent prediction.
- Evaluation on multiple COD benchmarks showing improved boundary fidelity and detection reliability.

## 2. Related work

Early COD methods primarily relied on CNN architectures to enhance structural perception through progressive refinement and feature aggregation. Multi-stage pipelines [28], graph-based reasoning [37], and rank-oriented supervision [15] have been explored to improve localization under weak foreground-background separability. Subsequent efforts placed greater emphasis on boundary sensitivity, incorporating contour-focused supervision [16], feature decomposition [8], and frequency-aware representations [3] to better capture subtle signal variations often concealed by camouflage. Despite these advances, most CNN-based designs remain largely dependent on deterministic feature learning, where structural cues are recovered indirectly through refinement rather than explicitly enforced during prediction.

Transformer-based models introduced stronger global reasoning capabilities, enabling the modelling of long-range dependencies. Uncertainty-guided attention has been explored to improve robustness in visually ambiguous regions [34], while collaborative pyramid modelling supports multi-scale token interaction [21]. Masked separable attention further enhances region isolation by limiting cross-token interference [36]. Recent work integrates frequency-spatial representations through entanglement transformer blocks [30] and incorporates external knowledge via text-guided interaction and adaptive data selection [33] to enhance discrimination. Despite improved semantic coherence, boundary ambiguity remains a persistent challenge, suggesting that global context alone does not fully resolve the structural uncertainty inherent to camouflaged scenes.

Generative formulations have recently reframed COD as a probabilistic generation task, allowing predictions to emerge through iterative refinement rather than deterministic decoding. Diffusion-based models exemplify this shift by progressively denoising mask representations [2]. Data-centric strategies further enhance robustness by synthesizing harder camouflaged training samples to expand supervision under challenging visual conditions [9]. In parallel, low-supervision paradigms explore noise-tolerant learning through noisy pseudo labels [38] and retrieval-based self-augmentation to stabilize training without ground-truth masks [4]. Nevertheless, structural constraints are rarely integrated explicitly into the generative process, leaving boundary formation dependent on emergent behaviour.

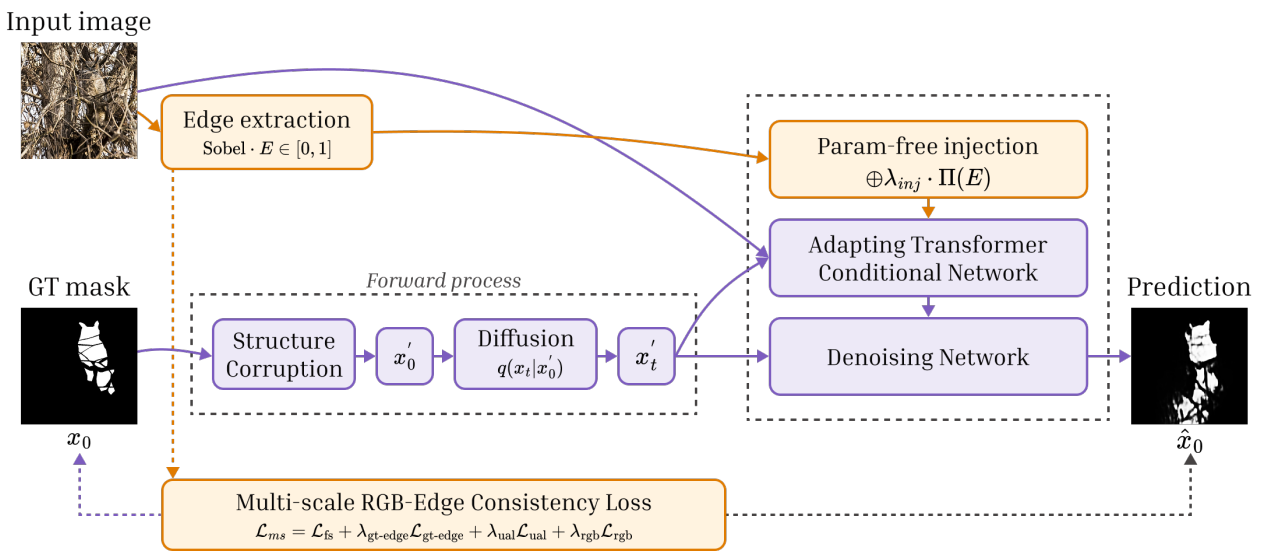

Figure 1. High-level overview of the proposed Bi-CamoDiffusion. Components highlighted in orange represent the contributions introduced in this work, while purple denotes the original CamoDiffusion components.

In summary, prior COD methods have improved mask prediction through feature refinement, attention mechanisms, and generative modelling. However, explicit structural constraints are still rarely integrated into the denoising process itself, leaving boundary formation largely dependent on emergent behaviour rather than on direct contour-aware guidance. The proposed method addresses this gap by coupling early edge-informed modulation with boundary-specific supervision in a diffusion-based COD pipeline.

## 3. Proposed approach

### 3.1. Overview

We build on CamoDiffusion [2], which formulates COD as conditional denoising diffusion for segmentation. Let $I$ denote an RGB image and $y$ be the ground-truth binary mask. During training, a forward noising (corruption) process gradually perturbs the clean target $y$ into a noisy variable $x_t \sim q(x_t \mid y, t)$ at diffusion step $t$. The diffusion model $f_\theta$ then learns to invert this process by predicting the clean mask variable $x_0$ from the noisy input $x_t$, conditioned on the image $I$ and the timestep $t$, so that $\hat{x}_0 = f_\theta(x_t, I, t)$.

The proposed model extends CamoDiffusion with a boundary-informed design that introduces no additional learnable parameters and remains compatible with pre-trained weights. The key idea is to exploit an RGB edge prior $E$ (computed once per image and provided during both training and inference) to encourage sharper contours without altering the backbone design. As illustrated in Fig. 1, $E$ is incorporated through two complementary components:

- **Boundary feature injection**, a parameter-free additive modulation applied to the earliest feature embedding stage, biases low-level representations toward boundary cues.

- **Boundary alignment loss**, an auxiliary supervision term that penalizes mismatch between the predicted mask gradients and the RGB edge prior.

These components sharpen contours and reduce boundary bleeding. Below, each component is described in detail.

### 3.2. Edge prior construction and feature injection

**Grayscale conversion** Given an RGB image $I \in \mathbb{R}^{3\times H\times W}$ with channels $I_R, I_G, I_B \in \mathbb{R}^{H\times W}$, first a single-channel luminance map via the standard ITU-R BT.601 weighted conversion is obtained, as in Eq. (1):

$$G = \mathcal{G}(I) = 0.299\, I_R + 0.587\, I_G + 0.114\, I_B, \quad G \in \mathbb{R}^{H\times W}. \tag{1}$$

This projection suppresses chromatic variation and retains only luminance structure, which is the dominant perceptual cue for spatial discontinuities. Crucially, camouflage patterns exploit colour and texture mimicry, but intensity transitions at object boundaries remain detectable even in challenging camouflaged scenes. Operating on $G$ rather than directly on $I$ thus reduces sensitivity to misleading chromatic gradients in the background.

**Sobel edge magnitude** The edge prior $E$ is obtained by applying the Sobel operator to $G$. The two directional gradient responses are computed via fixed $3\times 3$ convolution kernels shown in Eq. (2):

$$\mathbf{K}_x = \begin{pmatrix} -1 & 0 & 1 \\ -2 & 0 & 2 \\ -1 & 0 & 1 \end{pmatrix}, \qquad \mathbf{K}_y = \begin{pmatrix} -1 & -2 & -1 \\ 0 & 0 & 0 \\ 1 & 2 & 1 \end{pmatrix}, \tag{2}$$

yielding horizontal and vertical gradient maps $\partial_x G = \mathbf{K}_x * G$ and $\partial_y G = \mathbf{K}_y * G$, where $*$ denotes spatial convolution with unit-stride and same-padding. The edge magnitude is then defined as in Eq. (3):

$$E = \mathcal{S}(G) = \sqrt{(\partial_x G)^2 + (\partial_y G)^2 + \varepsilon}, \quad E \in [0,1], \tag{3}$$

where $\varepsilon > 0$ is a small constant for numerical stability. The result is clipped to $[0,1]$ to produce a normalised boundary map. The complete extraction pipeline is therefore established as shown in Eq. (4):

$$E = \mathcal{S}(\mathcal{G}(I)), \tag{4}$$

which is parameter-free and differentiable everywhere except at exact zero-gradient locations.

The backbone follows a hierarchical patch-embedding design, in which the first stage $\mathcal{C}_1$ operates at the highest spatial resolution and captures fine-grained structural detail. Subsequent stages progressively downsample the features and shift the representation toward higher-level semantics, thereby irreversibly discarding precise spatial support.

Injecting edge information at the earliest stage therefore ensures that boundary cues influence all subsequent feature transformations, rather than being introduced later as a corrective signal after spatial detail has already been lost. Furthermore, performing the injection at the earliest stage localises the intervention to a single, well-defined point, making the design transparent and easy to ablate.

**Parameter-free feature modulation** Let $F_1 = \mathcal{C}_1(I) \in \mathbb{R}^{B\times C\times H'\times W'}$ denote the feature map produced by the first convolutional stage, where $C$ is the embedding dimension and $H' = \lfloor H/s \rfloor$, $W' = \lfloor W/s \rfloor$ are the spatially downsampled dimensions with stride $s$. A modulated feature map $\tilde{F}_1$ is defined as in Eq. (5):

$$\tilde{F}_1 = F_1 + \lambda_{\text{inj}} \cdot \Pi(E), \tag{5}$$

where $\lambda_{\text{inj}} \in \mathbb{R}_{>0}$ is a fixed scalar hyperparameter and $\Pi : \mathbb{R}^{1\times H\times W} \to \mathbb{R}^{C\times H'\times W'}$ is a deterministic, parameter-free operator defined as in Eq. (6):

$$\Pi(E) = \mathbf{1}_C \otimes \phi(E), \tag{6}$$

with $\phi(E) \in \mathbb{R}^{1\times H' \times W'}$ denoting nearest-neighbour spatial rescaling of $E$ to the target resolution, and $\otimes$ denoting channel-wise broadcasting (i.e., replication across the $C$ dimension). No learnable projection is introduced; therefore, the modulation adds no trainable parameters.

The additive form in Eq. (5) is deliberate as an additive bias of this form can be interpreted as shifting the effective input to each downstream transformer block toward regions of high boundary activity, without distorting the feature distribution learned during pretraining. In contrast, approaches such as channel-wise gating or feature concatenation would either require new parameters or alter the input dimensionality seen by subsequent layers, both of which break compatibility with pretrained backbone weights.

**Boundary-sharpening pre-filter** The Sobel response $E$ measures gradient magnitude over a $3\times 3$ neighbourhood, which can yield spatially thick activations around boundaries. To obtain a sharper and more precisely localised boundary signal prior to injection, an fixed discrete Laplacian operator $\mathbf{K}_L$ to $E$ before channel expansion can be applied. The Laplacian kernel used is shown in Eq. (7):

$$\mathbf{K}_L = \begin{pmatrix} 0 & 1 & 0 \\ 1 & -4 & 1 \\ 0 & 1 & 0 \end{pmatrix}, \tag{7}$$

and the sharpened signal is shown in Eq. (8):

$$\phi(E) = \left| \mathbf{K}_L * \tilde{E} \right|, \tag{8}$$

where $\tilde{E}$ is $E$ rescaled to $H' \times W'$ and $|\cdot|$ denotes the element-wise absolute value, thereby retaining responses of both polarities. As a second-order derivative operator, the Laplacian responds maximally at zero-crossings of intensity transitions, which typically coincide with object boundaries, and produces thinner and more localised activations than the first-order Sobel response. Since $\mathbf{K}_L$ is a fixed kernel, this step introduces no additional parameters and does not affect backbone compatibility. The full modulated representation $\tilde{F}_1$ is then forwarded unchanged to the remaining stages of the backbone.

### 3.3. Multi-scale RGB-Edge Consistency Loss

While the parameter-free feature modulation introduced in this work biases the network representation toward boundary cues, it does not by itself guarantee that the predicted mask contours align with the edge structure of the input image. To address this limitation, a multi-scale RGB-Edge composite training objective is introduced, combining region-level segmentation accuracy with explicit boundary constraints derived directly from the input image. The objective is organised around four complementary terms, each targeting a distinct aspect of boundary quality in diffusion-based COD, and is jointly evaluated across multiple spatial scales. Each component is described below.

**Focal-structure loss** The base segmentation term must handle two concurrent challenges: hard pixels near camouflage boundaries where the model is most uncertain, and the structural bias toward under-segmenting thin or elongated regions. Both are addressed with a compound loss that fuses a boundary-weighted focal cross-entropy with a weighted Intersection over Union (IoU) term.

Let $\hat{y} = \sigma(z) \in [0,1]^{H\times W}$ denote the predicted probability map obtained by applying the sigmoid $\sigma$ to the logit output $z$, and let $y \in \{0,1\}^{H\times W}$ be the ground-truth mask. A spatially adaptive weight map that emphasises pixels in the vicinity of object boundaries is defined, as in Eq. (9):

$$w(y) = 1 + \alpha \left| \bar{y} - y \right|, \quad \bar{y} = \mathrm{AvgPool}_k(y), \tag{9}$$

where $\mathrm{AvgPool}_k$ denotes average pooling with kernel size $k = 31$ and same padding, and $\alpha = 5$. The difference $|\bar{y} - y|$ is large near boundary pixels and small in homogeneous interior or background regions, so $w(y)$ concentrates supervision where it is most needed.

The boundary-weighted focal cross-entropy is then defined in Eq. (10):

$$\mathcal{L}_{\text{fbce}}(z,y) = \frac{\sum_i w_i \cdot (1 - p_t^i)^{\gamma} \cdot \mathrm{BCE}(z_i, y_i)}{\sum_i w_i + \varepsilon}, \tag{10}$$

where $p_t^i = \hat{y}_i \cdot y_i + (1-\hat{y}_i)(1-y_i)$ is the probability assigned to the correct class at pixel $i$, $\gamma = 2$ is the focusing exponent, and $\mathrm{BCE}(\cdot,\cdot)$ denotes binary cross-entropy. The factor $(1 - p_t^i)^{\gamma}$ down-weights well-classified pixels and encourages the model to focus on the hard boundary pixels already emphasised by $w$.

The complementary weighted IoU term penalises region-level under-segmentation, as shown in Eq. (11):

$$\mathcal{L}_{\text{wIoU}}(z,y) = 1 - \frac{\sum_i w_i\, \hat{y}_i\, y_i + 1}{\sum_i w_i\,(\hat{y}_i + y_i) - \sum_i w_i\, \hat{y}_i\, y_i + 1}, \tag{11}$$

where the $+1$ Laplace smoothing prevents division by zero on empty predictions. The full focal-structure loss given by Eq. (12):

$$\mathcal{L}_{\text{fs}} = \mathcal{L}_{\text{fbce}} + \mathcal{L}_{\text{wIoU}}. \tag{12}$$

**Ground-truth edge loss** Even with the focal weighting above, the loss operates on per-pixel classification values and does not directly penalise errors in the spatial gradient of the prediction. A mask that is correct on average but blurred at the boundaries may incur a low $\mathcal{L}_{\text{fs}}$ yet fail precisely at object contours. This limitation is addressed by directly supervising the gradient field of the prediction. Let $\mathcal{S}(\cdot)$ denote the Sobel magnitude operator defined in Eq. (3). The ground-truth edge term is defined in Eq. (13):

$$\mathcal{L}_{\text{gt-edge}}(z,y) = \left\| \mathcal{S}(\hat{y}) - \mathcal{S}(y) \right\|_1, \tag{13}$$

which penalises the $\ell_1$ distance between the Sobel response of the predicted probability map and that of the binary ground-truth mask. This formulation is differentiable with respect to $z$ through $\hat{y} = \sigma(z)$ and the Sobel convolutions, allowing gradients to flow directly into the network from edge misalignment errors.

**Uncertainty-aware loss** Diffusion-based models involve intermediate denoising states $x_t$ that are inherently noisy and may produce predictions with spatially uneven confidence. Pixels for which the model remains uncertain, that is, where $\hat{y}_i \approx 0.5$, are penalised more strongly in order to discourage persistently ambiguous responses and promote sharper mask formation. Accordingly, the uncertainty map is defined as in Eq. (14):

$$u_i = 1 - \left| 2\hat{y}_i - 1 \right|^2 \in [0,1], \tag{14}$$

which attains its maximum value of 1 when $\hat{y}_i = 0.5$ (maximally uncertain) and is zero for fully confident predictions ($\hat{y}_i \in \{0,1\}$). The uncertainty-aware loss then uses this map as a spatially-varying penalty weight given by Eq. (15):

$$\mathcal{L}_{\text{ual}}(z,y) = \frac{1}{HW} \sum_i u_i, \tag{15}$$

encouraging the network to reduce spatial uncertainty and converge toward more decisive, sharply bounded predictions. This term is particularly relevant during early diffusion timesteps, when $x_t$ is highly corrupted and the model must produce meaningful predictions despite noisy intermediate states.

**RGB-Edge Consistency Loss** The three terms above rely exclusively on ground-truth mask annotations for boundary supervision. However, the mask annotation itself may be imprecise near ambiguous contours and carries no direct information about the image evidence supporting each predicted boundary. To address this limitation, an additional supervision signal derived directly from the input image, namely the RGB edge prior $E$ defined in Eq. (4), is introduced. Rather than enforcing pixel-wise agreement between the prediction and $E$, which would conflate all image gradients with object boundaries, the gradient field of the prediction is instead aligned with $E$ using Eq. (16):

$$\mathcal{L}_{\text{rgb}}(z, E) = \left\| \mathcal{S}(\hat{y}) - \tilde{E} \right\|_1, \tag{16}$$

where $\tilde{E}$ is a sanitised version of $E$ obtained by applying a light average-pool smoothing followed by threshold rescaling given by Eq. (17):

$$\tilde{E} = \left( \frac{\text{AvgPool}_3(E) - \tau}{1 - \tau} \right)_+, \tag{17}$$

with $\tau = 0.25$, where $(\cdot)_+ = \max(\cdot, 0)$ denotes rectification. The smoothing suppresses salt-and-pepper noise in the Sobel response, and the threshold $\tau$ discards weak gradients arising from background textures, retaining only strong, salient discontinuities as supervision targets. This term provides image-grounded boundary supervision that is independent of annotation quality, thereby complementing $\mathcal{L}_{\text{gt-edge}}$ with a signal derived from raw image evidence.

**Multi-scale aggregation and total objective** All four terms are jointly evaluated across a set of spatial scales $\mathcal{S} = \{1.0, 0.5, 0.25\}$ with corresponding normalised weights $\omega = \{1.0, 0.25, 0.125\}$. At each scale $s \in \mathcal{S}$, the logit map $z$, the ground-truth mask $y$, and the edge prior $E$ are bilinearly resampled to a resolution of $\lfloor sH \rfloor \times \lfloor sW \rfloor$. The scale-aggregated objective is defined in Eq. (18):

$$\begin{aligned} \mathcal{L}_{\text{ms}} = \frac{1}{\sum_s \omega_s} \sum_{s \in \mathcal{S}} \omega_s \big[ \mathcal{L}_{\text{fs}}^{(s)} + \lambda_{\text{gt-edge}} \mathcal{L}_{\text{gt-edge}}^{(s)} \\ + \lambda_{\text{ual}} \mathcal{L}_{\text{ual}}^{(s)} + \lambda_{\text{rgb}} \mathcal{L}_{\text{rgb}}^{(s)} \big], \end{aligned} \tag{18}$$

where $\lambda_{\text{gt-edge}} = 0.01$, $\lambda_{\text{ual}} = 0.01$, and $\lambda_{\text{rgb}} = 0.005$ are fixed scalar coefficients. The coarser scales ($s \in \{0.5, 0.25\} \subset \mathcal{S}$) receive lower weights to ensure that the full-resolution supervision remains dominant, while still providing a multi-scale gradient signal that stabilises training at early diffusion timesteps. The final training objective is defined in Eq. (19):

$$\mathcal{L}_{\text{total}} = \mathcal{L}_{\text{ms}}, \tag{19}$$

which unifies region accuracy, boundary sharpness, uncertainty minimisation, and image-grounded contour alignment into a single, fully differentiable objective.

# 4. Experiments

## 4.1. Datasets

The method is evaluated on three widely used benchmarks for COD: CAMO [12], COD10K [5], and NC4K [15].

CAMO focuses on difficult real-world conditions. It features high visual ambiguity, with 82% depicting animals and 62% exhibiting extreme color similarity between target and background. Structural complexity is prevalent, as over 50% of images contain heavy background clutter or intricate boundaries. Also, 35% of images involve small objects (under 10% of the image area) and 29% present partial occlusions, adding further scale and visibility challenges.

COD10K is a large-scale benchmark featuring camouflaged animals across 78 categories in natural environments. Unlike earlier datasets, it exhibits minimal center bias with objects distributed more uniformly. It presents extreme scale variation, ranging from 0.01% to 80.74% of the image area, reflecting a realistic distribution. Global and local contrast levels indicate that COD10K presents higher visual difficulty compared to other benchmarks.

NC4K spans diverse scenarios, from naturally camouflaged animals to artificial objects. It presents high domain variability, featuring significant shifts in image quality, illumination, textures, and camouflage patterns. This, along with background clutter and scale diversity, frequently expose model weaknesses in boundary delineation and low-conspicuity detection. It serves as a rigorous benchmark for assessing robustness and cross-domain generalization.

## 4.2. Experimental setup

Following common practices, the training/validation subsets of CAMO and COD10K are merged, yielding 7,066 images in total. For in-distribution evaluation, the results are reported on the official test splits of the CAMO test set (250 images) and the COD10K test set (2,026 images) separately. To assess cross-dataset generalization, the model is evaluated on NC4K, which is not used for training. Model performance is assessed using four COD standard metrics: the Structure-measure ($S_m$), the weighted F-measure ($F_\beta^w$), the E-measure ($E_m$), and the Mean Absolute Error (MAE).

For training, each RGB image and its corresponding ground-truth mask are resized to 416×416 pixels. The training operates with 64 noise dimensions and 30 denoising steps at inference, predicting directly in $x_0$ space with clipped denoising. Edge injection is configured with $\lambda_{\text{inj}} = 0.075$ and a Laplacian pre-filter, while the loss coefficients are set to $\lambda_{\text{gt-edge}} = 0.01$, $\lambda_{\text{ual}} = 0.01$, and $\lambda_{\text{rgb}} = 0.005$. The model is optimised with AdamW at a learning rate of $5 \times 10^{-5}$ with cosine annealing over 150 epochs, using a batch size of 32 on two NVIDIA A100 GPUs of 24GB each.

Table 1. Ablation of edge prior extraction operators on CAMO.

| **Edge method** | $S_m \uparrow$ | $E_m \uparrow$ | $F_\beta^w \uparrow$ | $MAE \downarrow$ |
|---|---|---|---|---|
| Prewitt | 0.8683 | 0.9314 | 0.8339 | 0.0448 |
| Laplacian | 0.8675 | 0.9267 | 0.8356 | 0.0483 |
| Canny | 0.8732 | 0.9335 | 0.8452 | 0.0439 |
| LoG | 0.8784 | 0.9410 | 0.8535 | 0.0416 |
| **Sobel** | **0.8971** | **0.9621** | **0.8740** | **0.0390** |

## 4.3. Results and analysis

**Results on CAMO** Table 1 compares different edge extraction operators on CAMO. Canny, despite being the most sophisticated operator, ranks third behind Sobel and Laplacian of Gaussian (LoG). Sobel consistently outperforms all alternatives across every metric, with a 3% $S_m$ gain and 0.009 MAE reduction over the weakest operator (Laplacian). This gap indicates that gradient smoothing before injection is the dominant factor, above operator complexity, which proves particularly effective under the complex structures present in CAMO.

Table 2 ablates the proposed loss components on CAMO. Adding $\mathcal{L}_{\text{gt-edge}}$ alone to the multi-scale baseline temporarily drops performance, suggesting that boundary supervision without uncertainty regularisation introduces conflicting gradients under CAMO's complex boundaries. $\mathcal{L}_{\text{ual}}$ alone recovers this drop, and the combination of both restores and improves over the baseline. The full objective, incorporating $\mathcal{L}_{\text{rgb}}$ as image-grounded boundary supervision, yields superior performance across evaluated metrics, with a 2.4% $S_m$ gain over the single-scale baseline, confirming that each component contributes to better optimising the model over the injected edge representations.

Fig. 2 complements the quantitative results, illustrating how the proposed edge injection and boundary-supervised loss consistently produce sharper and more complete masks compared to CamoDiffusion. The improvement is most evident in structurally challenging cases such as thin structures (a) and partially occluded objects (b), where the baseline either misses object parts or produces fragmented predictions, while our method recovers the full object extent more faithfully. Moreover, Table 3 shows that Bi-CamoDiffusion outperforms all compared methods on CAMO, confirming that the proposed edge injection and boundary-aware loss translate into consistent gains over the state-of-the-art (sota).

Table 2. Ablation of training losses under different combinations on CAMO dataset.

| Loss | Components | | | | | Performance | | | |
|---|---|---|---|---|---|---|---|---|---|
| | $\mathcal{L}_{\text{fs}}$ | MS | $\mathcal{L}_{\text{gt-edge}}$ | $\mathcal{L}_{\text{ual}}$ | $\mathcal{L}_{\text{rgb}}$ | $S_m \uparrow$ | $E_m \uparrow$ | $F_\beta^w \uparrow$ | MAE$\downarrow$ |
| $\mathcal{L}_{\text{fs}}$ (single-scale) | ✓ | | | | | 0.8728 | 0.9325 | 0.8392 | 0.0450 |
| $\mathcal{L}_{\text{fs}}$ (multi-scale) | ✓ | ✓ | | | | 0.8732 | 0.9314 | 0.8443 | 0.0456 |
| $\mathcal{L}_{\text{fs}} + \mathcal{L}_{\text{gt-edge}}$ | ✓ | ✓ | ✓ | | | 0.8650 | 0.9260 | 0.8321 | 0.0475 |
| $\mathcal{L}_{\text{fs}} + \mathcal{L}_{\text{ual}}$ | ✓ | ✓ | | ✓ | | 0.8721 | 0.9317 | 0.8459 | 0.0450 |
| $\mathcal{L}_{\text{fs}} + \mathcal{L}_{\text{gt-edge}} + \mathcal{L}_{\text{ual}}$ | ✓ | ✓ | ✓ | ✓ | | 0.8783 | 0.9368 | 0.8517 | 0.0415 |
| $\mathcal{L}_{\text{fs}} + \mathcal{L}_{\text{gt-edge}} + \mathcal{L}_{\text{ual}} + \mathcal{L}_{\text{rgb}}$ | ✓ | ✓ | ✓ | ✓ | ✓ | **0.8971** | **0.9621** | **0.8740** | **0.0390** |

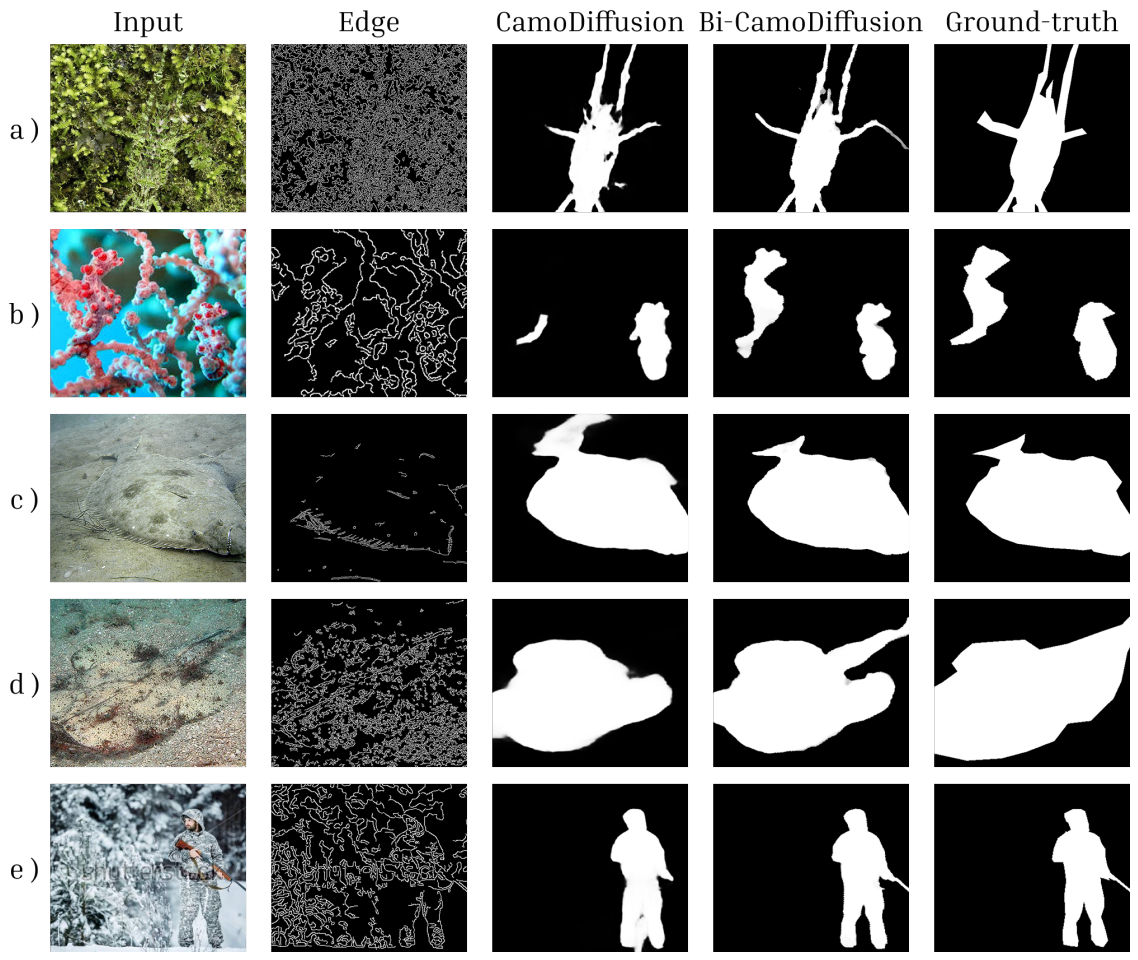

Figure 2. Qualitative comparison between our boundary-informed method and the baseline on the CAMO dataset.

**Results on COD10K** Table 4 reports the edge operators ablation on COD10K. Sobel performs best across metrics, showing a trend largely consistent with CAMO. Notably, the performance spread across operators is narrower than on CAMO, with a 1.7% $S_m$ gap between Sobel and the weakest operator, suggesting that on COD10K's highly variable object scales the choice of edge operator has less impact than in CAMO's structurally complex scenes.

Table 3. Performance comparison of our approach with other methods from the state-of-the-art on CAMO dataset.

| **Method** | $S_m \uparrow$ | $E_m \uparrow$ | $F_\beta^w \uparrow$ | $MAE \downarrow$ |
|---|---|---|---|---|
| EGNet [39] | 0.7320 | 0.7996 | 0.6036 | 0.1095 |
| PraNet [6] | 0.7692 | 0.8245 | 0.6625 | 0.0942 |
| F$^3$Net [31] | 0.7113 | 0.7407 | 0.5636 | 0.1087 |
| MINet [19] | 0.7480 | 0.7911 | 0.6370 | 0.0903 |
| SINet [5] | 0.7454 | 0.8035 | 0.6443 | 0.0915 |
| C$^2$F-Net [28] | 0.7961 | 0.8537 | 0.7187 | 0.0799 |
| PFNet [18] | 0.7823 | 0.8410 | 0.6952 | 0.0849 |
| MGL [37] | 0.7755 | 0.8160 | 0.6728 | 0.0884 |
| UGTR [34] | 0.7839 | 0.8215 | 0.6836 | 0.0863 |
| SINet-v2 [7] | 0.8201 | 0.8817 | 0.7426 | 0.0705 |
| BASNet [22] | 0.7943 | 0.8511 | 0.7174 | 0.0786 |
| OCENet [14] | 0.8019 | 0.8518 | 0.7234 | 0.0804 |
| BGNet [29] | 0.8116 | 0.8698 | 0.7485 | 0.0734 |
| ZoomNet [20] | 0.8197 | 0.8770 | 0.7522 | 0.0659 |
| CamoDiffusion [2] | 0.8659 | 0.9238 | 0.8263 | 0.0480 |
| FSEL [30] | 0.8850 | 0.9420 | 0.8500 | 0.0400 |
| **Bi-CamoDiffusion** | **0.8971** | **0.9621** | **0.8740** | **0.0390** |

Table 4. Ablation of edge prior extraction operators on COD10K.

| Edge method | $S_m \uparrow$ | $E_m \uparrow$ | $F_\beta^w \uparrow$ | $MAE \downarrow$ |
|---|---|---|---|---|
| Prewitt | 0.8647 | 0.9280 | 0.7848 | 0.0231 |
| Laplacian | 0.8720 | 0.9303 | 0.7967 | 0.0230 |
| Canny | 0.8693 | 0.9373 | 0.8103 | 0.0228 |
| LoG | 0.8781 | 0.9338 | 0.8053 | 0.0230 |
| **Sobel** | **0.8813** | **0.9491** | **0.8343** | **0.0200** |

Table 5 reports the loss ablation on COD10K. The pattern mirrors CAMO in that $\mathcal{L}_{\text{gt-edge}}$ alone temporarily hurts performance. A notable finding is that on COD10K, where objects span extreme scale variation, $\mathcal{L}_{\text{ual}}$ proves particularly impactful, recovering and surpassing the baseline even without $\mathcal{L}_{\text{gt-edge}}$, suggesting that uncertainty regularisation is especially beneficial when the model must handle highly variable object sizes. The full objective again yields superior results across all metrics, with $\mathcal{L}_{\text{rgb}}$ providing the final gain that consolidates all boundary supervision signals.

Likewise, Fig. 3 shows qualitative results on COD10K, where the proposed method more accurately delineates fragmented structures (c) and objects with complex, multi-pronged boundaries (b, d), while CamoDiffusion struggles to recover the full structural detail. Similarly, Table 6 confirms that Bi-CamoDiffusion outperforms existing methods on COD10K, achieving notable gains over ZoomNet, BGNet and CamoDiffusion in $S_m$, $E_m$ and $F_\beta^w$. FSEL, despite ranking second and outperforming all other competing methods, does not reach the levels achieved by Bi-CamoDiffusion, with $S_m$ and $E_m$ below 0.88 and 0.93 respectively, reflecting weaker structural understanding and contour alignment on this benchmark.

**Results on NC4K** Table 7 compares edge extraction operators on NC4K. Sobel outperforms others across metrics, with a 2% $S_m$ gain over Prewitt and the lowest MAE overall, showing that Sobel-extracted edge priors provide the most effective boundary cues for our model to generalise to unseen domains. Despite a competitive MAE, Canny edges yield the weakest $F_\beta^w$, suggesting that they are less informative for detection under NC4K's high variability.

Also, Table 8 shows the loss ablation on NC4K. The full objective is the only configuration to surpass 0.90 $S_m$ and 0.95 in $E_m$, indicating that the complete loss enables the model to capture the structural composition of camouflaged objects more accurately, which no partial combination achieves in this unseen domain. The remaining metrics also improve progressively, confirming that the boundary supervision introduced by each term transfers beyond the training distribution.

Fig. 4 shows qualitative results on NC4K. More diffi-culty is observed in generating accurate masks in this unseen domain, yet the proposed method consistently produces fewer false positive regions (b, d and e), with notably better delineation of small and elongated details, as evident in (e). Finally, Table 9 confirms this against the sota, with Bi-CamoDiffusion being the only method to surpass 0.90 in $S_m$, 0.95 in $E_m$, and achieve a MAE below 0.03, demonstrating superior structural understanding and contour precision even under the domain shift conditions of NC4K.

Table 5. Ablation of training losses under different combinations on COD10K dataset.

| Loss | Components | | | | | Performance | | | |
|---|---|---|---|---|---|---|---|---|---|
| | $\mathcal{L}_{\text{fs}}$ | MS | $\mathcal{L}_{\text{gt-edge}}$ | $\mathcal{L}_{\text{ual}}$ | $\mathcal{L}_{\text{rgb}}$ | $S_m \uparrow$ | $E_m \uparrow$ | $F_\beta^w \uparrow$ | MAE$\downarrow$ |
| $\mathcal{L}_{\text{fs}}$ (single-scale) | ✓ | | | | | 0.8715 | 0.9307 | 0.7918 | 0.0219 |
| $\mathcal{L}_{\text{fs}}$ (multi-scale) | ✓ | ✓ | | | | 0.8715 | 0.9332 | 0.7918 | 0.0218 |
| $\mathcal{L}_{\text{fs}} + \mathcal{L}_{\text{gt-edge}}$ | ✓ | ✓ | ✓ | | | 0.8686 | 0.9260 | 0.8004 | 0.0224 |
| $\mathcal{L}_{\text{fs}} + \mathcal{L}_{\text{ual}}$ | ✓ | ✓ | | ✓ | | 0.8745 | 0.9314 | 0.8102 | 0.0210 |
| $\mathcal{L}_{\text{fs}} + \mathcal{L}_{\text{gt-edge}} + \mathcal{L}_{\text{ual}}$ | ✓ | ✓ | ✓ | ✓ | | 0.8734 | 0.9367 | 0.8240 | 0.0209 |
| $\mathcal{L}_{\text{fs}} + \mathcal{L}_{\text{gt-edge}} + \mathcal{L}_{\text{ual}} + \mathcal{L}_{\text{rgb}}$ | ✓ | ✓ | ✓ | ✓ | ✓ | **0.8813** | **0.9491** | **0.8343** | **0.0200** |

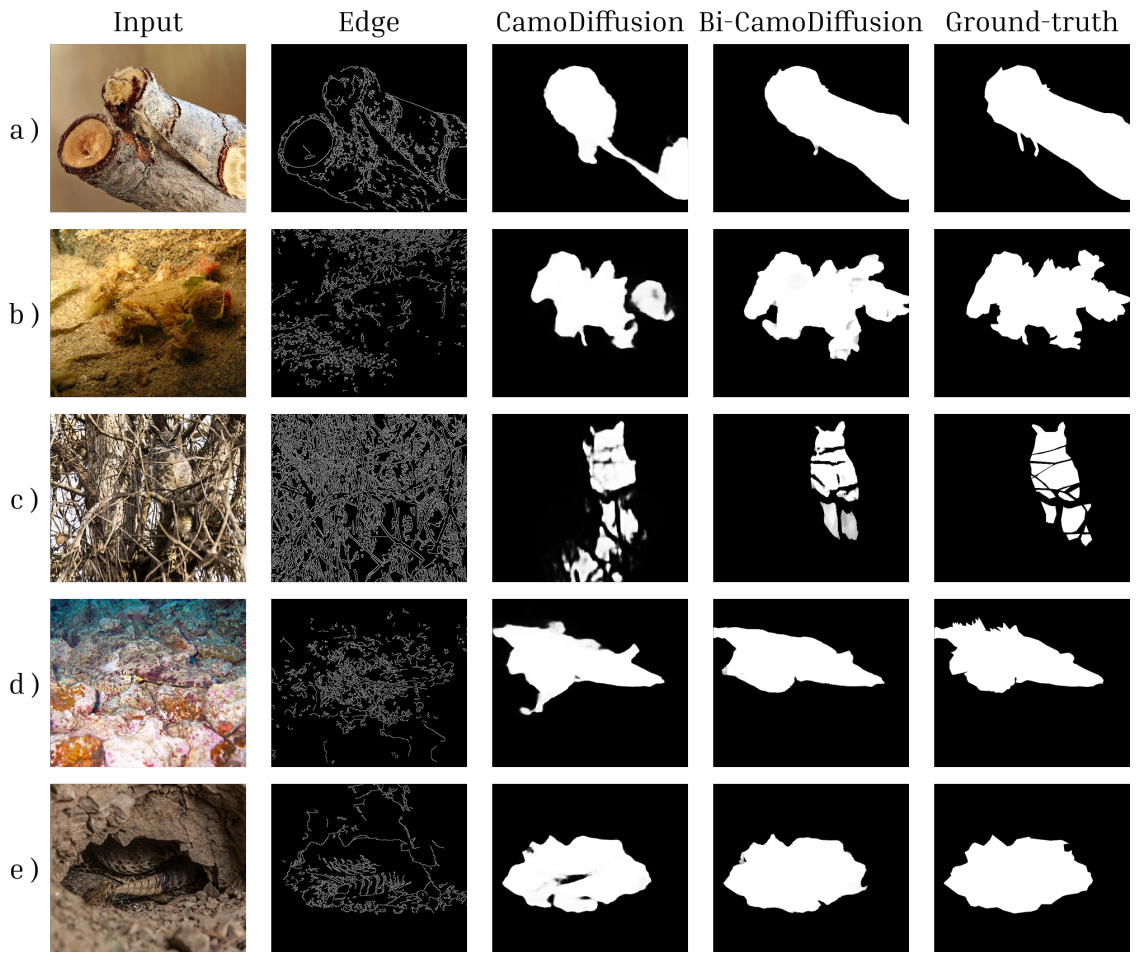

Figure 3. Qualitative comparison between our boundary-informed method and the baseline on the COD10K dataset.

Table 6. Performance comparison of our approach with other methods from the state-of-the-art on COD10K dataset.

| Method | $S_m \uparrow$ | $E_m \uparrow$ | $F_\beta^w \uparrow$ | $MAE \downarrow$ |
|---|---|---|---|---|
| EGNet [39] | 0.7365 | 0.8097 | 0.5174 | 0.0607 |
| PraNet [6] | 0.7894 | 0.8606 | 0.6294 | 0.0451 |
| F$^3$Net [31] | 0.7386 | 0.7951 | 0.5438 | 0.0513 |
| MINet [19] | 0.7697 | 0.8320 | 0.6085 | 0.0417 |
| SINet [5] | 0.7794 | 0.8642 | 0.6310 | 0.0426 |
| C$^2$F-Net [28] | 0.8130 | 0.8902 | 0.6862 | 0.0360 |
| PFNet [18] | 0.7998 | 0.8772 | 0.6599 | 0.0396 |
| MGL [37] | 0.8139 | 0.8513 | 0.6657 | 0.0353 |
| UGTR [34] | 0.8171 | 0.8519 | 0.6656 | 0.0356 |
| SINet-v2 [7] | 0.8151 | 0.8870 | 0.6796 | 0.0368 |
| BASNet [22] | 0.8176 | 0.8905 | 0.6990 | 0.0342 |
| OCENet [14] | 0.8272 | 0.8935 | 0.7071 | 0.0327 |
| BGNet [29] | 0.8307 | 0.9007 | 0.7219 | 0.0326 |
| ZoomNet [20] | 0.8384 | 0.8876 | 0.7288 | 0.0289 |
| CamoDiffusion [2] | 0.8599 | 0.9274 | 0.7732 | 0.0245 |
| FSEL [30] | 0.8730 | 0.9280 | 0.8000 | 0.0210 |
| **Bi-CamoDiffusion** | **0.8813** | **0.9491** | **0.8343** | **0.0200** |

Table 7. Ablation of edge prior extraction operators on NC4K.

| Edge method | $S_m \uparrow$ | $E_m \uparrow$ | $F_\beta^w \uparrow$ | $MAE \downarrow$ |
|---|---|---|---|---|
| Prewitt | 0.8842 | 0.9337 | 0.8246 | 0.0321 |
| Laplacian | 0.8910 | 0.9173 | 0.8102 | 0.0314 |
| Canny | 0.8975 | 0.9256 | 0.8001 | 0.0310 |
| LoG | 0.9001 | 0.9307 | 0.8234 | 0.0345 |
| **Sobel** | **0.9043** | **0.9532** | **0.8517** | **0.0209** |

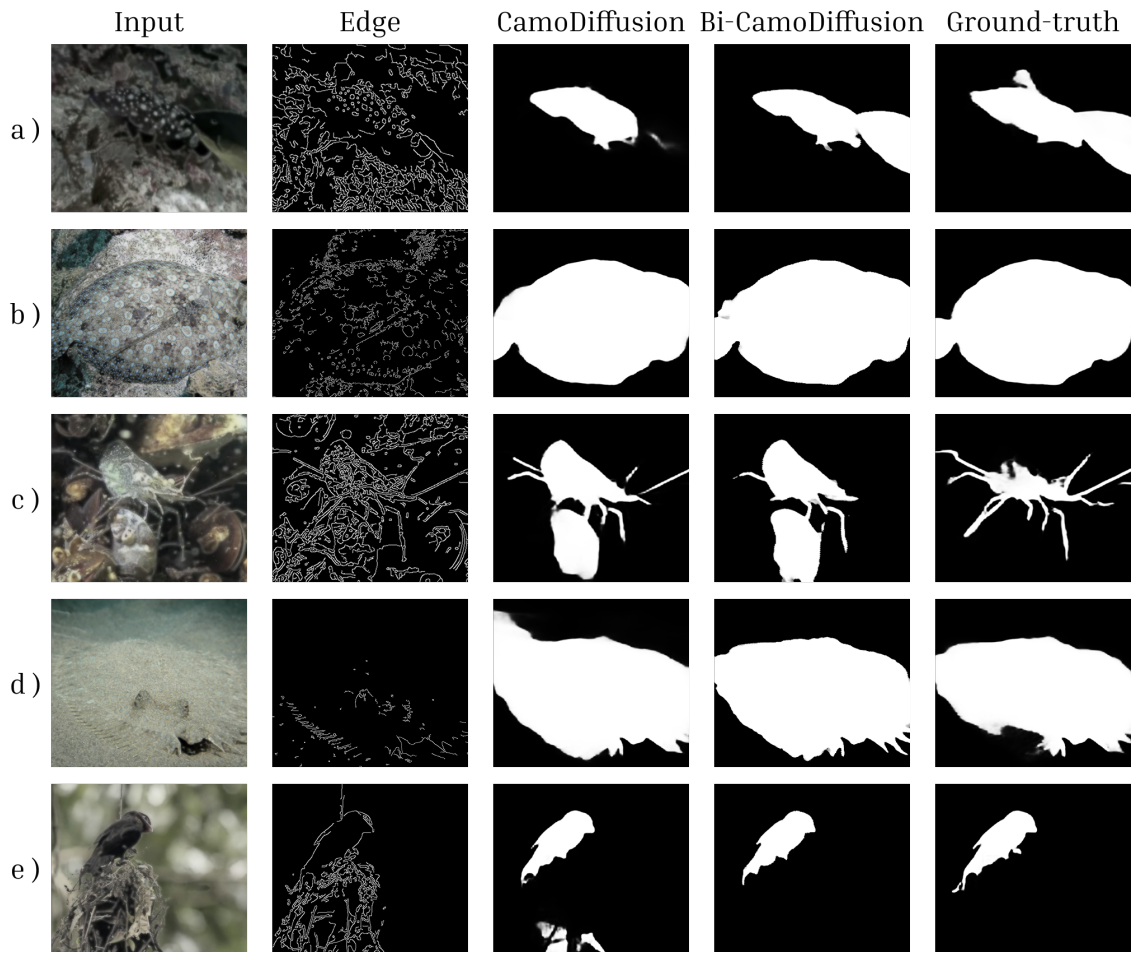

Figure 4. Qualitative comparison between our boundary-informed method and the baseline on the NC4K dataset.

# 5. Limitations and future work

Despite evaluating Bi-CamoDiffusion on multiple datasets and against different competing methods, some limitations of this study must be acknowledged. First, our evaluation follows the official splits provided by each benchmark. While this ensures comparability with prior work, it does not include targeted analyses of specific COD conditions, such as naturally and artificially camouflaged cases separately. Future work could expand the evaluation to specific scenarios of interest. Likewise, a natural direction for future work is to test alternative methods for extracting the edge prior. This study primary relied on classical image-processing operators; therefore, more advanced alternatives can be explored, including deep learning-based edge or contour detectors that may provide stronger and more semantically consistent boundary cues. Another promising direction is to extend the model design beyond a fixed, parameter-free prior injection by incorporating priors through lightweight, learnable integration modules and analyzing the impact of such an approach.

Table 8. Ablation of training losses under different combinations on NC4K dataset.

| Loss | Components | | | | | Performance | | | |
|---|---|---|---|---|---|---|---|---|---|
| | $\mathcal{L}_{fs}$ | MS | $\mathcal{L}_{gt\text{-}edge}$ | $\mathcal{L}_{ual}$ | $\mathcal{L}_{rgb}$ | $S_m \uparrow$ | $E_m \uparrow$ | $F_\beta^w \uparrow$ | MAE$\downarrow$ |
| $\mathcal{L}_{fs}$ (single-scale) | ✓ | | | | | 0.8874 | 0.9365 | 0.8225 | 0.0319 |
| $\mathcal{L}_{fs}$ (multi-scale) | ✓ | ✓ | | | | 0.8880 | 0.9389 | 0.8250 | 0.0302 |
| $\mathcal{L}_{fs} + \mathcal{L}_{gt\text{-}edge}$ | ✓ | ✓ | ✓ | | | 0.8890 | 0.9377 | 0.8129 | 0.0300 |
| $\mathcal{L}_{fs} + \mathcal{L}_{ual}$ | ✓ | ✓ | | ✓ | | 0.8889 | 0.9282 | 0.8269 | 0.0263 |
| $\mathcal{L}_{fs} + \mathcal{L}_{gt\text{-}edge} + \mathcal{L}_{ual}$ | ✓ | ✓ | ✓ | ✓ | | 0.8899 | 0.9383 | 0.8358 | 0.0297 |
| $\mathcal{L}_{fs} + \mathcal{L}_{gt\text{-}edge} + \mathcal{L}_{ual} + \mathcal{L}_{rgb}$ | ✓ | ✓ | ✓ | ✓ | ✓ | **0.9043** | **0.9532** | **0.8517** | **0.0209** |

Table 9. Performance comparison of our approach with other methods from the state-of-the-art on NC4K dataset.

| **Method** | $S_m \uparrow$ | $E_m \uparrow$ | $F_\beta^w \uparrow$ | $MAE \downarrow$ |
|---|---|---|---|---|
| EGNet [39] | 0.7771 | 0.8408 | 0.6386 | 0.0751 |
| PraNet [6] | 0.8222 | 0.8761 | 0.7245 | 0.0588 |
| F$^3$Net [31] | 0.7800 | 0.8244 | 0.6556 | 0.0695 |
| MINet [19] | 0.8122 | 0.8618 | 0.7195 | 0.0555 |
| SINet [5] | 0.8080 | 0.8173 | 0.7227 | 0.0576 |
| C$^2$F-Net [28] | 0.8383 | 0.8974 | 0.7624 | 0.0490 |
| PFNet [18] | 0.8290 | 0.8874 | 0.7453 | 0.0527 |
| MGL [37] | 0.8326 | 0.8666 | 0.7392 | 0.0526 |
| UGTR [34] | 0.8394 | 0.8744 | 0.7463 | 0.0519 |
| SINet-v2 [7] | 0.8472 | 0.9027 | 0.7698 | 0.0476 |
| BASNet [22] | 0.8414 | 0.8968 | 0.7708 | 0.0479 |
| OCENet [14] | 0.8533 | 0.9025 | 0.7846 | 0.0450 |
| BGNet [29] | 0.8510 | 0.9067 | 0.7884 | 0.0444 |
| ZoomNet [20] | 0.8528 | 0.8957 | 0.7844 | 0.0434 |
| CamoDiffusion [2] | 0.8833 | 0.9343 | 0.8372 | 0.0325 |
| FSEL [30] | 0.8920 | 0.9410 | **0.8530** | 0.0300 |
| Bi-CamoDiffusion | **0.9043** | **0.9532** | 0.8517 | **0.0209** |

# 6. Conclusions

This work introduces Bi-CamoDiffusion, a boundary-informed diffusion model for Camouflaged Object Detection. The original CamoDiffusion is extended by incorporating a parameter-free prior injection process that modulates early-stage embeddings with edge-aware information. Furthermore, a loss function is proposed that combines spatial precision, structural edge constraints, and uncertainty-aware supervision into a unified objective to ensure the model captures both global semantics and local boundary details. Experiments on CAMO, COD10K, and NC4K datasets demonstrate that Bi-CamoDiffusion achieves a significant improvement in object delineation, producing finer detections and a notable reduction in false positives. Also, our model consistently outperforms existing state-of-the-art methods, demonstrating superior boundary precision and a deeper structural understanding.

# Acknowledgement

This material is based upon work supported by the Air Force Office of Scientific Research under award number FA9550-24-1-0206; and partially supported by the Grant PID2021-128945NB-I00 funded by MICIU/AEI/ 10.13039/501100011033 and by ERDF/EU and Grant PID2024-162815NB-I00 funded by MICIU/AEI/ 10.13039/501100011033 and by ERDF/EU. The authors acknowledge the support of the Generalitat de Catalunya CERCA Program to CVC's general activities.